\documentclass[11pt,letterpaper]{article}
\usepackage[utf8]{inputenc}
\usepackage{CJKutf8}
\usepackage{fullpage}
\usepackage{amsfonts,eucal,amsbsy,amsopn,amsmath}
\usepackage{url}
\usepackage{enumerate}
\usepackage[sort&compress]{natbib}
\usepackage{fancyhdr}
\usepackage{sectsty}
\usepackage[dvipsnames,usenames]{color}
\definecolor{orange}{rgb}{1,0.5,0}
\usepackage{caption}
\usepackage{tikz}
\usetikzlibrary{shapes,arrows}
\usepackage{multirow}

\allsectionsfont{\normalsize}
\newcommand{\ignore}[1]{}

\newcounter{glosscounter}
\newcommand{\gloss}{\stepcounter{glosscounter}(\theglosscounter)}

\newcommand{\attribute}[1]{{\texttt{+\textsc{#1}}}}
\newcommand{\textttsc}[1]{{\texttt{\textsc{#1}}}}
\newcommand{\romaji}[1]{{\emph{#1}}}
\newcommand{\japanese}[1]{{\begin{CJK}{UTF8}{min}#1\end{CJK}}}
\newcommand{\japromaji}[2]{{\romaji{#1} (\japanese{#2})}}

\title{A Morphological Analyzer for Japanese Nouns, Verbs and Adjectives\thanks{This morphological analyzer is done as part of the project requirements for the Spring 2013 NLP Lab (11-712) at Carnegie Mellon University (\protect\url{http://www.cs.cmu.edu/~nasmith/NLPLab/}).}}
\author{Yanchuan Sim\\Language Technologies Institute\\Carnegie Mellon University\\Pittsburgh, PA 15213\\\texttt{yanchuan@cmu.edu}}
\date{April 26, 2013}

\begin{document}
\maketitle
\begin{abstract}
We present an open source morphological analyzer for Japanese nouns, verbs and adjectives.\footnote{The source code and tool is available at \url{https://bitbucket.org/skylander/yc-nlplab/}.}
The system builds upon the morphological analyzing capabilities of MeCab \citep{matsumoto1999japanese} to incorporate finer details of classification such as politeness, tense, mood and voice attributes.
We implemented our analyzer in the form of a finite state transducer using the open source finite state compiler FOMA toolkit \citep{hulden2009foma}.
\end{abstract}

\section{Basic Information about Japanese}

The Japanese language is spoken by more than 100 million speakers (mostly in Japan).
It is an agglutinative language with a SOV word order.
The Japanese writing system makes extensive use of Chinese characters, also known as \romaji{kanji}, along with scripts, \romaji{hiragana} and \romaji{katakana}, which are syllabic.

There are 3 main lexical classes in Japanese that exhibit morphology, they are the nouns, verbs and adjectives.
In Japanese, adverbs are often adjectives suffixed with a special morpheme.
As such, it is often not considered a separate class of words.
In \S\ref{sec:survey}, we will discuss the phenomenon for each of these word classes.

\section{Past Work on the Japanese morphology}

In traditional Japanese morphological analysis, a lexicon is assumed. The lexicon is a list of pairs of a word and its corresponding part-of-speech. As Japanese is an unsegmented language, past work on Japanese language morphology analyzers require the use of a segmenter. Often, the segmentation step is conducted jointly with the morphological analysis step in a rule based \citep{matsumoto1991user} or machine learning framework \citep{matsumoto1999japanese,kudo2004applying}.

The current state of the art Japanese language part of speech and morphological analyzer is MeCab\footnote{\url{http://mecab.googlecode.com/svn/trunk/mecab/doc/index.html}} \citep{matsumoto1999japanese}, which is an extended from ChaSen\footnote{\url{http://chasen-legacy.sourceforge.jp/}} \citep{matsumoto1991user}, but using CRFs\citep{lafferty2001conditional} instead of HMMs to model the morpheme sequences.

MeCab segments words and provides annotations for 69 different part of speech categories\footnote{\url{http://mecab.googlecode.com/svn/trunk/mecab/doc/posid.html}}.
These categories, although comprehensive, do not cover the certain morphological phenomena in detail, such as the differences between honorifics used with Japanese nouns, and the finer grammatical attributes of verbal and adjectival conjugations.
In this lab, we will seek to improve upon current morphological analysis tools by filling in these missing gaps.

\section{Available Resources}

We will be using MeCab as a preprocessing tool for word segmentation and POS tagging. MeCab achieved $F_\beta$ scores of above 0.96 on standard Japanese text corpora that has been segmented and annotated with POS tags.

We gathered the following text corpora:
\begin{enumerate}[(i)]
\item Tanaka corpus \citep{tanaka2001compilation} contains 149,909 English-Japanese sentence pairs from various sources such as books, songs, bible, etc.
These sentences are word segmented and contain \romaji{kana} readings for ambiguous Kanji characters.
However, they do not contain morphological annotations.

\item Japanese-Multilingual Dictionary \citep{breen2004jmdict} is a multilingual lexical database with Japanese as the pivot language.
It is updated almost everyday and contains numerous information about lexical items, such as \romaji{kanji}/\romaji{kana}/\romaji{romaji} readings, part of speech, conjugation group and sense tags.
We used the dictionary for our verbs and adjective lexicon.
\end{enumerate}

Table~\ref{tab:corpus-lex-stats} shows the distribution of different lexical types in the above two corpora.

\begin{table}[ht]
\centering
\begin{tabular}{|l|c|c|c|}
\hline
\multirow{2}{*}{Lexical type} & \multicolumn{2}{|c|}{Tanaka corpus} & Japanese-Multilingual dictionary\\ \cline{2-4}
& Tokens & Types & Types\\ \hline
Nouns & 331,577 & 15,769 & -\\ \hline
Verbs & 247,544 & 4,588 & 40,507\\ \hline
Adjectival verbs & 29,869 & 584 & 4,032\\ \hline
Adjectival nouns & 22,138 & 1113 & 10,242\\ \hline
\end{tabular}
\caption{Statistics of lexical items in Tanaka corpus and Japanese-Multilingual dictionary.\label{tab:corpus-lex-stats}}
\end{table}

\section{Survey of Phenomena in Japanese}\label{sec:survey}

\subsection{Nouns}
Nouns are usually not inflected.
However, the Japanese language do have an extensive \emph{honorific} system, and nouns often take politeness prefixes and suffixes (for animate objects), a small group of suffixes are also used to refer to collective nouns.

\paragraph{Honorific prefixes} The prefixes \japromaji{o-}{お-} and \japromaji{go-}{ご-} are prefixes applied to nouns (and sometimes verbs) to convey a respectful or polite tone of speech.
Native Japanese words are preceded by \romaji{o-} while Sino-Japanese words are usually preceded by \romaji{go-}, although there are exceptions to both cases.
\begin{figure}[ht]
\centering
\begin{minipage}[b]{0.45\linewidth}
\begin{tabular}{lll}
\gloss & \japanese{お} & \japanese{金}\\
& \romaji{o} & \romaji{kane}\\
& \textttsc{hon} & money\\
& \multicolumn{2}{l}{`money'}
\end{tabular}
\end{minipage}
\begin{minipage}[b]{0.45\linewidth}
\begin{tabular}{lll}
\gloss & \japanese{ご} & \japanese{飯}\\
& \romaji{go} & \romaji{han}\\
& \textttsc{hon} & rice\\
& \multicolumn{2}{l}{`rice'}
\end{tabular}
\end{minipage}
\caption{Examples of honorific prefixes.}
\end{figure}

\paragraph{Honorific suffixes} Honorific suffixes are almost always attached to person names and occasionally other non-human objects (such as animals, inanimate objects, etc).
These suffixes depend on the formality of the dialogue, the speaker's familiarity and status difference with the referred person, as well as the gender/age of the referred person.

\begin{figure}[ht]
\centering
\begin{minipage}[b]{0.3\linewidth}
\begin{tabular}{lll}
\gloss & \japanese{田中} & \japanese{さん}\\
& \romaji{tanaka} & \romaji{san}\\
& Tanaka & \textttsc{hon}\\
& \multicolumn{2}{l}{`Mr. Tanaka'}
\end{tabular}
\end{minipage}
\begin{minipage}[b]{0.3\linewidth}
\begin{tabular}{lll}
\gloss & \japanese{田中} & \japanese{君}\\
& \romaji{tanaka} & \romaji{kun}\\
& Tanaka & \textttsc{hon}\\
& \multicolumn{2}{l}{`Tanaka' (informal)}
\end{tabular}
\end{minipage}
\begin{minipage}[b]{0.3\linewidth}
\begin{tabular}{lll}
\gloss & \japanese{女王} & \japanese{様}\\
& \romaji{joou} & \romaji{sama}\\
& Queen & \textttsc{hon}\\
& \multicolumn{2}{l}{`Your majesty'}
\end{tabular}
\end{minipage}
\caption{Examples of honorific suffixes.}
\end{figure}

\paragraph{Collective suffixes} Collective suffixes are appended to nouns to signify a collective group of the noun.
These suffixes are often used with singular pronouns to refer to its plural counterpart.
Common collective suffixes are \emph{-tachi} and \emph{-ra}.
Occasionally, these suffixes maybe used with proper animate nouns to signify a group including the mentioned noun (collective, not pluralizing).
For instance, \emph{Tanaka-tachi} maybe translated as `Tanaka and his friends' or `Tanaka and his family' but not `some people named Tanaka'.
We take this distinction into account when collective suffixes are used in the context of pronouns.
\begin{figure}[ht]
\centering
\begin{minipage}[b]{0.33\linewidth}
\begin{tabular}{lll}
\gloss & \japanese{私} & \japanese{達}\\
& watashi & tachi\\
& \textttsc{1per} & \textttsc{pl}\\
& \multicolumn{2}{l}{`We'}
\end{tabular}
\end{minipage}
\begin{minipage}[b]{0.3\linewidth}
\begin{tabular}{lll}
\gloss & \japanese{僕} & \japanese{等}\\
& boku & ra\\
& \textttsc{1per} & \textttsc{pl}\\
& \multicolumn{2}{l}{`We' (informal)}
\end{tabular}
\end{minipage}
\begin{minipage}[b]{0.35\linewidth}
\begin{tabular}{lll}
\gloss & \japanese{田中} & \japanese{達}\\
& \romaji{tanaka} & \romaji{tachi}\\
& Tanaka & \textttsc{col}\\
& \multicolumn{2}{l}{`Tanaka and his friends'}
\end{tabular}
\end{minipage}
\caption{Examples of collective suffixes.}
\end{figure}

\paragraph{Pronouns}
Japanese is a pro-drop language, and hence are used less frequently than in English.
Hence, pronouns are usually treated morphologically like nouns and plural pronouns are formed with collective suffixes.
In addition, pronouns are considered an open lexical class (since it behaves just like a noun), which means that new pronouns are being ``created'' to be used in specialized contexts (age, familiarity, different levels of formality, etc).
As such, we will only consider pronouns that are found in our corpus.

\subsection{Verbs}
Verbs in the Japanese language are conjugated extensively depending on their grammatical categories.
Due to the agglutinative nature of the language, the suffixes are fairly productive and in casual speech, many contractions are used.
Generally, verbs can be conjugated for aspect, polarity, politeness, mood, voice, tense.
Details about verb conjugations will be described in \S\ref{sec:verb-phenomena}.

\subsection{Adjectives}\label{sec:adj-overview}
Japanese do not have adjectives in a syntactic sense.
Unlike western languages where adjectives are usually inflected similar to noun declensions, Japanese adjectives are conjugated like a verb.
This is because words that behave semantically as adjectives are technically considered as specialized verbs (\romaji{i}-adjectives) or nominals (\romaji{na}-adjectives).
There are several other classes of adjectives, which are more commonly used in Classical Japanese, and in Modern Japanese are limited to a select few special cases.
There are two main classes of adjectives: adjectival verbs and adjectival nouns.

\section{System Design}

\subsection{High level overview}

Our morphological analyzer will focus on the task of analyzing the morphology of individual tokens.
Hence, we will design our analyzer to take input from the processed output of MeCab, a high quality state of the art Japanese part of speech tagger and morphological analyzer.
MeCab is a CRF classifier that segments and tags each tokens with a part of speech category.
It has a hierarchical tag set (of up to 4 levels), starting with 8 coarse categories and 68 fine categories in total.
In the case of nouns, it provides a coarse morphological analysis by segmenting the morphemes for honorifics and collectivism and classifying it as \emph{suffixes} or \emph{prefixes} belonging to \emph{persons, locations, quantifiers}, etc.
Hence, our contribution would be to incorporate finer details of classification such as politeness and gender attributes.
Details of our implementation for nouns would be described in \S\ref{sec:noun-impl-details}.

For verbs, adjectives and adverbs, the morphemes are not segmented as the surface forms undergo various morphophonological changes.
Thus, the analysis by MeCab is limited to the forms of the verbs and its lemma.
Our analyzer will therefore focus on extracting the morphological attributes such as tense, mood, voice, etc.
The details of our implementation for verbs will be discussed in \S\ref{sec:verb-impl-details}.

\begin{figure}[ht]
  \centering
  \tikzstyle{block} = [rectangle, draw, text width=3cm, text centered, rounded corners, minimum height=4em]
  \tikzstyle{cloud} = [draw, ellipse, node distance=3cm, minimum height=2em]
  \tikzstyle{line} = [draw, -latex']
  \begin{tikzpicture}
    \node [cloud] (data) {sentence};
    \node [block, right of=data, node distance=3.5cm] (mecab) {MeCab segmentation and tagging};
    \node [block, right of=mecab, node distance=6.5cm] (analyzer) {Morphological analyzer};
    \node [cloud, right of=analyzer, node distance=3.5cm] (output) {Output};

    \path [line] (data) -- (mecab);
    \path [line] (mecab) -- node [above, text width=4cm, text centered] {\small relevant tokens and tag information} (analyzer);
    \path [line] (analyzer) -- (output);
\end{tikzpicture}
\caption{Raw sentences will be fed to MeCab for segmentation and POS tagging. After which, the individual tokens, along with the POS and lemma information will be fed to our FST-based morphological analyzer for analysis.}
\label{fig:flowchart}
\end{figure}
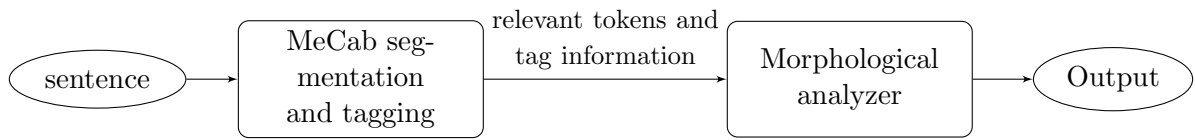

Figure \ref{fig:flowchart} shows the high level view of our morphological analyzer pipeline.

\subsection{Morphological analyzer system details}

We implement our analyzer by building finite state transducers to process and generate finer analyses.
We will use Foma\footnote{https://code.google.com/p/foma/}, which is a popular open source tool compiler, programming language, and C library for constructing finite-state automata and transducers for NLP applications.

Our morphological analyzer will consist of two independent FSTs, for handling (i) nouns, and (ii)~verbs/adjectives, which we will describe in the sequel.

\subsection{FSTs for analyzing nouns}\label{sec:noun-impl-details}

Japanese nouns have fairly regular conjugations, as such, we will design our FST as a transducer that takes MeCab's part of speech information as input, and translate it into its analysis.

From MeCab's output, we will have a post-processing script that parses MeCab's output and convert it into a format suitable for our analyzers.
We will do this using a Python script and the input to the FST would be
\begin{quote}
\begin{verbatim}
N#<token1>/<pos_id1> <token2>/<pos_id2> ...$
\end{verbatim}
\end{quote}

For some nouns, the prefixes and suffixes have been segmented into separate tokens.
Thus, an input may consist of several tokens with different part of speech IDs.

\paragraph{Noun morphology phenomena}
The following noun morphology are handled by our analyzer:
\begin{enumerate}
\item Honorific prefixes: \japanese{お-, ご-} (common) and \japanese{御-} (usually only found in literary writings).
Output will be marked with a \attribute{polite} attribute.

\item Formal/informal honorific suffixes for animate objects: \japanese{-ちゃん, -さん, -君} and \japanese{-様}.
For \japanese{-ちゃん} and \japanese{-君}, the output will be marked with \attribute{informal}, while \japanese{-様} will be marked with \attribute{formal}.
The very common \japanese{-さん} suffix is not tagged formal/informal because the use of it is standard in Japanese (as a default polite form) and hence its use do not convey much politeness information beyond that of typical social scenarios.
However, all objects with honorific suffixes mentioned above will have an \attribute{animate} attribute.

\item Collective suffixes: \japanese{達, 等, 方, たち, ら} and \japanese{かた}.
Output will be marked with a \attribute{collective} attribute.

\item Table~\ref{tab:pronouns} presents a list of pronouns that are handled by our analyzer and their morphological attributes.
Furthermore, consecutive sequences of pronouns-collectives are handled as a single entity even if MeCab segments them into separate tokens (i.e \attribute{plural} instead of \attribute{collective}).

\begin{table}[ht]
\centering
\begin{tabular}{|l|c|c|c|c|}
\hline
\multicolumn{1}{|c|}{Pronoun} & Person & Number & Gender & Formality\\ \hline
\multicolumn{5}{|c|}{First person} \\ \hline
\japanese{私, わたし} & \attribute{1per} & \attribute{sg} & &\\ \hline
\japanese{我, 吾, 余} (archaic) & \attribute{1per} & \attribute{sg} & & \attribute{formal}\\ \hline
\japanese{こちら} & \attribute{1per} & \attribute{sg} & & \attribute{informal}\\ \hline
\japanese{儂, わし} & \attribute{1per} & \attribute{sg} & \attribute{male} &\\ \hline
\japanese{己, おのれ} & \attribute{1per} & \attribute{sg} & \attribute{male} & \attribute{formal}\\ \hline
\japanese{僕}  & \attribute{1per} & \attribute{sg} & \attribute{male} & \attribute{informal}\\ \hline
\japanese{あたし, うち} & \attribute{1per} & \attribute{sg} & \attribute{female} & \attribute{informal}\\ \hline
\japanese{われわれ, 我々} & \attribute{1per} & \attribute{pl} & & \attribute{informal}\\ \hline
\japanese{僕ら, 僕達} & \attribute{1per} & \attribute{pl} & \attribute{male} & \attribute{informal}\\ \hline
\multicolumn{5}{|c|}{Second person} \\ \hline
\japanese{あなた, 貴方} & \attribute{2per} & \attribute{sg} & &\\ \hline
\japanese{あんた, 君} & \attribute{2per} & \attribute{sg} & & \attribute{informal}\\ \hline
\japanese{きさま, お前} & \attribute{2per} & \attribute{sg} & \attribute{male} & \attribute{informal}\\ \hline
\japanese{君たち} & \attribute{2per} & \attribute{pl} & & \attribute{informal}\\ \hline
\multicolumn{5}{|c|}{Third person} \\ \hline
\japanese{かれ, やつ, 奴} & \attribute{3per} & \attribute{sg} & & \attribute{informal}\\ \hline
\japanese{彼女} & \attribute{3per} & \attribute{sg} & \attribute{female} & \attribute{informal}\\ \hline
\japanese{奴ら, 奴等, 彼ら} & \attribute{3per} & \attribute{pl} & & \attribute{informal}\\ \hline
\japanese{彼女ら} & \attribute{3per} & \attribute{pl} & \attribute{female} & \attribute{informal}\\ \hline
\end{tabular}
\caption{List of Japanese pronouns handled by our system and their morphological analysis.}
\label{tab:pronouns}
\end{table}

\item Noun possessive marker \japromaji{no}{の}. Noun phrases ending with a \romaji{no}\footnote{MeCab tags such tokens as \japanese{名詞,非自立,一般}, with a POS ID of 63.} token are replaced with a \attribute{possessive} attribute.

\end{enumerate}

\section{First System Analysis (Nouns)}

We extracted all the types that were tagged as nouns in Tanaka corpus.
Table~\ref{tab:corpus-lex-stats} show statistics about nouns in the corpus.
We randomly selected 1,000 types to manually evaluate and analyze for mistakes.

Table~\ref{tab:noun-eval-stats} shows some statistics about the nouns evaluation set.
In general, most nouns in the corpus does not exhibit any morphology.
For those that do contain morphological elements, our analyzer was able to identify them with high accuracy.
However, our analyzer made a few mistakes (10 on the evaluation set to be exact), which can all be attributed to segmentation/tagging errors returned by MeCab, which was unable to segment the sentences correctly.

\begin{table}[ht]
\centering
\begin{tabular}{|l|c|}
\hline
Noun types (total) & 1000\\ \hline
Analysis accuracy & .990\\ \hline
Examples containing morphology & 117\\ \hline
\end{tabular}
\caption{Evaluation results for nouns.\label{tab:noun-eval-stats}}
\end{table}

\subsection{Examples}

Table~\ref{tab:noun-analysis-examples} shows example analyses produced by our noun FST.

\begin{table}[ht]
\centering
\begin{tabular}{|l|l|}
\hline
\multicolumn{1}{|c|}{Input to FST} & \multicolumn{1}{|c|}{Analysis}\\ \hline
\multirow{2}{*}{\japanese{\texttt{N\#お/30 医者/38 様/55\$}}} & \japanese{医者}\attribute{formal}\attribute{animate}\attribute{polite}\\
& `doctor'\\ \hline
\multirow{2}{*}{\japanese{\texttt{N\#少年/38 達/51\$}}} & \japanese{少年}\attribute{collective}\\
& `youth'\\ \hline
\multirow{2}{*}{\japanese{\texttt{N\#彼女/59 達/51\$}}} & \textttsc{prn}\attribute{3per}\attribute{female}\attribute{pl}\attribute{informal}\\
& `the gals'\\ \hline
\end{tabular}
\caption{Example analyses produced by noun FST.\label{tab:noun-analysis-examples}}
\end{table}

\section{FSTs for analyzing verbs and adjectives}\label{sec:verb-impl-details}

Japanese verbs and adjectives are handled very similarly.
In fact, most adjectives can be considered a form of specialized verbs, and feature very similar conjugation rules.
As such, we shall use a single FST to handle these two categories of words.

\subsection{Verb morphology phenomena}\label{sec:verb-phenomena}

Lexically, almost every verb in Japanese is belongs to one of four conjugation groups:
\begin{enumerate}[(i)]
\item Ichidan verb (\japanese{一段動詞}),
\item Godan verb (\japanese{五段動詞}),
\item Sa row irregular conjugations (\japanese{サ行変格活用}), and
\item Ka row irregular conjugations (\japanese{カ行変格活用})
\end{enumerate}

A verb in Japanese is made up two parts, a stem form and a series of morphemes.
The stem form is usually written in \emph{kanji} along with a \emph{kana} character that denote one of the 6 verb forms in Japanese.
The morphemes suffixes are usually written in \emph{kana}.
We refer the user to a comprehensive grammar guidebook for more information about the stem forms.
Here, we will briefly describe the morpheme suffixes that are handled by our analyzer.
We will now describe the suffixes that are analyzed by our system:

\begin{enumerate}
\item Dictionary forms: The root form of a Japanese verb is known as the dictionary form.
When used in its dictionary form, it denotes an imperfective aspect and informal tone.
The dictionary form is returned by our analyzer as a verb's lemma.

\item Perfective aspects (\attribute{pfv}) are usually denoted by \japromaji{-ta}{-た} or \japromaji{-da}{-だ} suffix, but various phonetic changes are made depending on a verb's last syllable.

\item Negation (\attribute{neg}) are usually denoted by \japromaji{-nai}{-ない} suffix.
However, in casual and colloquial Japanese, many different suffixes for negation are encountered.
Some examples that we found in our corpus are \japanese{-な, -ぬ, -ん, -ず, -ずな, -なきゃ, -せぬ, -ざる}.

\item Passive form (\attribute{pasv}) have a \japromaji{-reru}{-れる} suffix which can be treated as an \romaji{ichidan} verb and conjugated.

\item \romaji{Te}-forms (\attribute{te}), also known as conjunctive forms, allow multiple verbs to be used together.
Example (9) and (10) shows some \romaji{te}-form verbs being used.
The verbs that conjoin with the \romaji{te}-form stems can be considered a form of helper auxiliary verb, of which there are many in Japanese.
Also, the newly conjoined verb is usually conjugated as an \romaji{ichidan} verb.
Since there are to many possible auxiliaries for us to consider, we only consider the auxiliary \japromaji{-iru}{-いる} when used with \romaji{te}-forms, which denotes the progressive aspect (\attribute{prog}).
Other auxiliaries would be analyzed independently, and it would be up to the downstream application to take into account the presence of a preceding \attribute{te} form token.
It would be easy to extend the FST to handle other helper auxiliary verbs.

\begin{figure}[ht]
\centering
\begin{minipage}[b]{0.45\linewidth}
\begin{tabular}{lll}
\gloss & \japanese{寝て} & \japanese{いる}\\
& \romaji{nete} & \romaji{iru}\\
& sleep\attribute{te} & to be\\
& \multicolumn{2}{l}{`to be sleeping'}
\end{tabular}
\end{minipage}
\begin{minipage}[b]{0.45\linewidth}
\begin{tabular}{lll}
\gloss & \japanese{食べて} & \japanese{おく}\\
& \romaji{tabete} & \romaji{oku}\\
& eat\attribute{te} & put/place\\
& \multicolumn{2}{l}{`to eat in advance'}
\end{tabular}
\end{minipage}
\caption{Examples of \romaji{te}-form conjunctive verbs.}
\end{figure}

In addition, there are also many auxiliaries that compound with non \romaji{te}-form stems.

\item Conditional moods (\attribute{cond}) are used in conditionals and denoted (usually) by the \japromaji{-ba}{-ば} and \japromaji{-ra}{-ら} suffixes.

\item Volitional form (\attribute{vol}) expresses intention and are used with the \japromaji{-u}{-う} suffix.

\item Imperative form (\attribute{imp}) are characterized by verbs conjugating to have an \romaji{e} phonetic ending.

\item Causative forms (\attribute{caus}) generally have a \japromaji{-seru}{-せる} / \japromaji{-saru}{-さる} suffix.

\item Potential forms (\attribute{pot}) are used to express that one has the ability to do something.

\item Like nouns, verbs have a polite form (\attribute{pol}) that is usually characterized by a \japromaji{-masu}{-ます} suffix.
\end{enumerate}

\subsection{Phonological rules}

Like most languages, there are several phonological rules that apply to the conjugated verbs and adjectives.
One such common and regular rule is phonological change when ending with a \textsf{[t]} consonant.
For example \romaji{te}-form verb endings \romaji{-ite, -chite, -rite} change to a double consonant, \romaji{-tte},
endings in \romaji{-bite, -mite, -nite} change to \romaji{-nde} (voicing of the \romaji{te} syllable),
endings in \romaji{-kite} change to \romaji{-ite}, and
endings in \romaji{-gite} change to \romaji{-ide} (voicing of the \romaji{te} syllable).
The above double consonants and voicing patterns also applies when dealing with perfective aspect endings \romaji{-ta}.

There are also numerous contractions that are used in colloquial conversations.
For instance, the progressive marker \romaji{-teiru} becomes \romaji{-teru} and \romaji{-teshimau} becomes \romaji{-chau}.

\subsection{FST as a generator}
Due to the agglutinative nature of the language, the morphemes suffixes can be added in many different arbitrary order.
Hence, we have designed the FST as a generator, meaning the output of the FST is the surface form of a conjugated verb and the input is its lemma and morphological feature.
The FST inherently has no knowledge of valid or invalid orderings of these suffixes and thus tend to over generate surface forms for a particular input.

When the FST is being used, it is trivial to reverse the direction of the FST, and return the underlying form of the verb token.
However, it is very possible for a verb token to have multiple analysis and it is up to the downstream application to decide between possible analyses.

\subsection{Adjective morphology phenomena}\label{sec:adj-phenomena}.

As described in \S\ref{sec:adj-overview}, there are two main classes of adjectives, and we describe how our analyzer handle them here.

\paragraph{Adjectival verbs (\romaji{i}-adjectives)}
end with \japromaji{-i}{-い} and can be conjugated with almost all the suffixes for verbs.
These adjectives are also often found in predicate positions, denoting their verb-like nature.
For instance, when a verb takes on a \romaji{-nai} negation suffix, it can be treated morphologically like an i-adjective, and likewise, when an \romaji{i}-adjective is conjugated for \romaji{te}-form, it behaves just like any other te-form verb.

\paragraph{Adjectival nouns (\romaji{na}-adjectives)} always occur with a form of the copula \japromaji{-da}{-だ}.
Because adjectival nouns are usually formed from nouns, it has led many linguists to consider them a type of nominal.
The copula (considered an irregular \romaji{ichidan} verb) can conjugated just like any other verb.
It is thus treated as such by our analyzers.

\paragraph{Other adjective classes} have limited use in modern Japanese.
In our analyzer, we have chosen to ignore these adjectives and not analyze them.

\section{Second System Analysis (verbs and adjectives)}

We extracted all the tokens that were tagged as verbs or adjectives in the Tanaka corpus.
Table~\ref{tab:corpus-lex-stats} show statistics about the verbal and adjectival lexical items in the corpus.
From all the tokens, we extracted 2,000 verb and 1,000 adjective tokens for manual evaluation.
Since for each token, the FST produce multiple analyses (there can also be multiple valid analyses absent context), we measure the precision/recall performance of our analyzer.
Precision is computed by
\[
\frac{1}{|\text{evaluation set}|}\sum_{i\in\text{evaluation set}}\frac{\text{\# of correct analyses}}{\text{\# of analyses}}
\]
and recall is computed by
\[
\frac{1}{|\text{evaluation set}|}\sum_{i\in\text{evaluation set}} \mathbb{I}(\text{\# of correct analyses} > 0)
\]
Table~\ref{tab:vbadj-eval-stats} presents the results of our evaluation.
Since our FST over generates (see table~\ref{tab:verb-analysis-examples} and \ref{tab:adj-analysis-examples} for such examples), we were able to obtain reasonably good recall.
However, we missed many items due to a lack of coverage in our lexicon.
The Japanese-Multilingual dictionary, though comprehensive, does not provide coverage of all the lexical items that appeared in the Tanaka corpus.
This problem is more pronounced in verbs due to verbs being a much larger and more productive lexical class.

The precision for our verb analysis were quite low, which we attribute to the agglutinative nature of the language.
The morpheme suffixes are a productive class, and the Tanaka corpus contain a significant amount of conversational text and colloquial language.
On the other hand, adjectives tend to exhibit far lesser conjugations and hence we are able to achieve better precision.
In fact, 77.4\% of the surface forms in the adjective evaluation set were found to be in dictionary form (i.e were not conjugated).
Adjectives are usually conjugated when they are used in the predicate position.

\begin{table}[ht]
\centering
\begin{tabular}{|l|c|c|}
\hline
Lexical type & Precision & Recall\\ \hline
Verbs & 0.776 & 0.892\\ \hline
Adjectives & 0.892 & 0.990\\ \hline
\end{tabular}
\caption{Evaluation results for verbs and adjectives.\label{tab:vbadj-eval-stats}}
\end{table}

\subsection{Examples}

See table~\ref{tab:verb-analysis-examples} and \ref{tab:adj-analysis-examples} for verb and adjective example analyses respectively.

\begin{table}[ht]
\centering
\begin{tabular}{|l|l|}
\hline
\multicolumn{1}{|c|}{Input to FST} & \multicolumn{1}{|c|}{Analysis}\\ \hline
\multirow{2}{*}{\japanese{言った}} & \japanese{言う}\attribute{v}\attribute{pfv}\\
& `said'\\ \hline
\multirow{2}{*}{\japanese{助かりでした}} & \japanese{助かる}\attribute{v}\attribute{pol}\attribute{pfv}\\
& `saved'\\ \hline
\multirow{3}{*}{\japanese{信じられている}} & \japanese{信じる}\attribute{v}\attribute{pasv}\attribute{te}\attribute{prog}\\
& \japanese{信ずる}\attribute{v}\attribute{pasv}\attribute{te}\attribute{prog} (less common)\\
& `had been believing'\\ \hline
\multirow{3}{*}{\japanese{見られない}} & \japanese{見る}\attribute{v}\attribute{pot}\attribute{neg}\\
& \japanese{見る}\attribute{v}\attribute{pasv}\attribute{neg}\\
& `not able to see' or `was not seen'\\ \hline
\end{tabular}
\caption{Example verbal analyses produced by verbs/adjectives FST.\label{tab:verb-analysis-examples}}
\end{table}

\begin{table}[ht]
\centering
\begin{tabular}{|l|l|}
\hline
\multicolumn{1}{|c|}{Input to FST} & \multicolumn{1}{|c|}{Analysis}\\ \hline
\multirow{2}{*}{\japanese{容易な}} & \japanese{容易}\attribute{adj}\attribute{adv}\\
& `easily'\\ \hline
\multirow{2}{*}{\japanese{悪くない}} & \japanese{悪い}\attribute{adj}\attribute{neg}\\
& `not bad'\\ \hline
\multirow{2}{*}{\japanese{奇麗だった}} & \japanese{奇麗}\attribute{adj}\attribute{pfv}\\
& `was pretty'\\ \hline
\multirow{3}{*}{\japanese{好きです}} & \japanese{好く}\attribute{v}\attribute{pol}\\
& \japanese{好き}\attribute{adj}\attribute{pol}\\
& `like' (verb) or `is lovely' (adjective)\\ \hline
\multirow{2}{*}{\japanese{美しかった}} & \japanese{美しい}\attribute{adj}\attribute{pfv}\newline\\
& `was beautiful'\\ \hline
\end{tabular}
\caption{Example adjectival analyses produced by verbs/adjectives FST.\label{tab:adj-analysis-examples}}
\end{table}

\section{Final Revisions}

Many of the errorneous analyses produced can be eliminated by means of simple heuristics.
We built a post-processing recognizer (instead of an FST) that takes as input: (i) the dictionary forms (lemmas) returned by MeCab, and (ii) each of the various analyses returned by our; and decide whether the analysis
\begin{enumerate}[(i)]
  \item contained invalid orderings of morphemes, or
  \item matches MeCab's dictionary forms.
\end{enumerate}
We believe this will help reduce the number of analyses generated per input, thus improving precision without hurting recall.

\section{Future Work}
For future work, we propose the following improvements
\begin{enumerate}[(i)]
  \item increasing coverage of verbal and adjectival lexical items,
  \item extending coverage of analyzer to handle \romaji{te}-form conjunctive verbs,
  \item extending coverage of adjective analyzer to handle rare adjectives and verb classes, and
  \item possibly more heuristics in the post-processing to reduce over-generation.
\end{enumerate}

\bibliographystyle{plainnat}
\bibliography{references}
\label{lastpage}
\end{document}